\pdfoutput=1

\documentclass[11pt]{article}

\usepackage[]{ACL2023}

\usepackage{times}
\usepackage{latexsym}

\usepackage{soul}
\usepackage[T1]{fontenc}

\usepackage[utf8]{inputenc}

\usepackage{microtype}

\usepackage{inconsolata}
\usepackage{multirow}
\usepackage{booktabs}
\usepackage{graphicx}
\usepackage{float}
\usepackage{algorithm}
\usepackage{algpseudocode}
\usepackage{caption}
\usepackage{subfigure}
\usepackage{amssymb}

\newcommand\model{UGDRE}
\newcommand\blfootnote[1]{%
  \begingroup
  \renewcommand\thefootnote{}\footnote{#1}%
  \addtocounter{footnote}{-1}%
  \endgroup
}

\title{Uncertainty Guided Label Denoising for Document-level Distant \\ Relation Extraction}

\author{Qi Sun$^{1,2}$, Kun Huang$^{1}$, Xiaocui Yang$^{2,3}$, Pengfei Hong$^{2}$, \\ 
    \textbf{Kun Zhang}$^{1}$$^{*}$, \textbf{and} \textbf{Soujanya Poria}$^{2}$$^{*}$ \\
  $^{1}$Nanjing University of Science and Technology \\
  $^{2}$Singapore University of Technology and Design,
  $^{3}$Northeastern University \\ 
  \texttt{\{319106003718, huangkun, zhangkun\}@njust.edu.cn}, \\
  \texttt{\{pengfei\_hong, sporia\}@sutd.edu.sg}, \\
  \texttt{yangxiaocui@stumail.neu.edu.cn}
  }

\begin{document}
\maketitle

\begin{abstract}
Document-level\blfootnote{$^*$Corresponding author} relation extraction (DocRE) aims to infer complex semantic relations among entities in a document. Distant supervision (DS) is able to generate massive auto-labeled data, which can improve DocRE performance. Recent works leverage pseudo labels generated by the pre-denoising model to reduce noise in DS data. However, unreliable pseudo labels bring new noise, e.g., adding false pseudo labels and losing correct DS labels. Therefore, how to select effective pseudo labels to denoise DS data is still a challenge in document-level distant relation extraction. To tackle this issue, we introduce uncertainty estimation technology to determine whether pseudo labels can be trusted. In this work, we propose a \textbf{D}ocument-level distant \textbf{R}elation \textbf{E}xtraction framework with \textbf{U}ncertainty \textbf{G}uided label denoising, \model. Specifically, we propose a novel instance-level uncertainty estimation method, which measures the reliability of the pseudo labels with overlapping relations. By further considering the long-tail problem, we design dynamic uncertainty thresholds for different types of relations to filter high-uncertainty pseudo labels. We conduct experiments on two public datasets. Our framework outperforms strong baselines by 1.91 $F_1$ and 2.28 Ign $F_1$ on the RE-DocRED dataset. \footnote{https://github.com/QiSun123/UGDRE}
\end{abstract}

\section{Introduction}
Document-level Relation Extraction (DocRE) aims to extract relations among entities in a document. In contrast to the conventional RE task that mainly focuses on sentence-level \citep{zhou2016attention,guo2019attention,tian2021dependency}, DocRE is more challenging due to the complex semantic scenarios, discourse structure of the document, and long-distant interactions between entities. 

To understand complex inter-sentence entity relations, most existing methods employ transformer-based \citep{huang2021three,zhou2021atlop,li2021mrn} or graph-based models \cite{nan2020reasoning,zeng2020double,zeng2021sire} that aggregate effective entity representations. Although these methods achieve reasonable performance, they heavily rely on the large-scale human-annotated corpus, which is time-consuming and labor-intensive. Distant supervision mechanism \citep{mintz2009distant} provides large-scale distantly supervised (DS) data auto-labeled by existing relational triples from knowledge bases \citep{xie2021revisiting}. Recent works observe that leveraging DS data to pretrain DocRE models can improve performance by a great margin \cite{xu2021entity,zhou2022none,wang2022unified}.

\begin{figure}
	\centering
	\includegraphics[width=0.29\textheight]{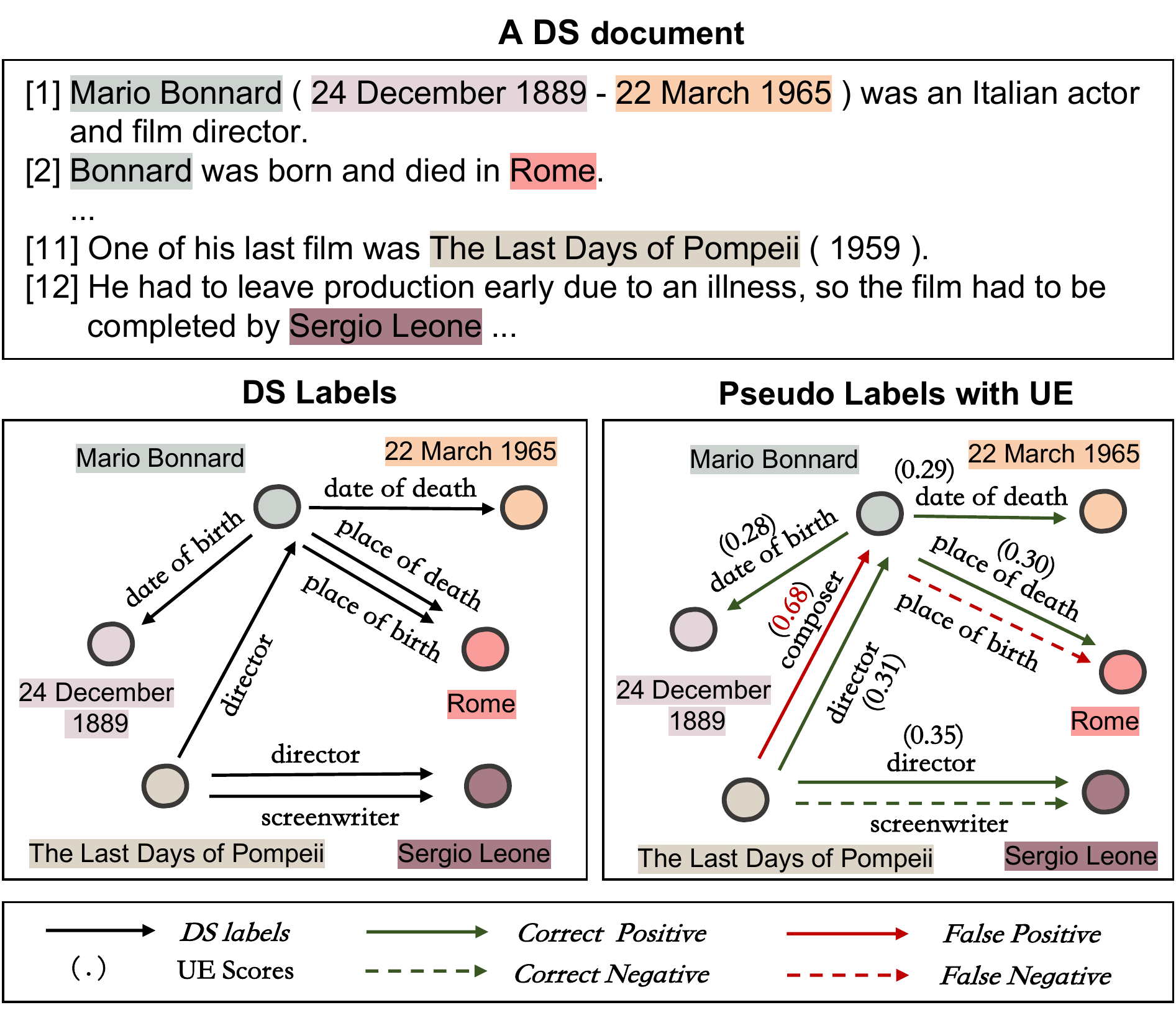}
	\caption{An example of the DS document. We present two types of noise caused by pseudo labels. One is adding new false labels as shown by the solid red line. Another is losing the correct DS label as shown by the red dotted line. We also show the proposed instance-level uncertainty estimation (UE) scores of pseudo labels. We present partly entities that are marked with different colors.}
	\label{fig:1}  
 
\end{figure}

Despite a great quantity of training data auto-labeled by distant supervision can enhance the performance of the model, noisy labels in DS data are non-negligible. \citet{yao2019docred} show that there are 61.8\% noisy inter-sentence instances in their provided document-level distant relation extraction dataset. Current efforts \citep{xiao2020denoising, tan2022KDDocRED} to alleviate the noise problem mainly employ a pre-denoising model. They train a DocRE model on human-annotated data first and then re-label DS data by the trained model.

However, the above methods still persist the risk of noise induction in the DS data due to false positive re-labeling. Besides, false negative pseudo labels also lead to the loss of effective labels in DS data. As shown in Figure \ref{fig:1}, we obtain an extra false instance \textit{(The Last Days of Pompeii, Mario Bonnard, composer)} and lose the correct DS instance \textit{(Mario Bonnard
, Rome, place of birth)}, when merely relying on pseudo labels. Thus, how to mitigate noise caused by pseudo labels and take full advantage of DS data is still a challenge in document-level distant RE.

In this work, we propose a \textbf{D}ocument-level distant \textbf{R}elation \textbf{E}xtraction framework with \textbf{U}ncertainty \textbf{G}uided label denoising, \model. 
We first train a pre-denoising DocRE model with both DS and human-annotated data to generate pseudo labels. Since false pseudo labels predicted by the pre-denoising model are inevitable, we introduce Uncertainty Estimation (UE) to determine whether model predictions can be trusted or not. As shown in Figure \ref{fig:1}, we can remove the false positive pseudo instance \textit{(The Last Days of Pompeii, Mario Bonnard, composer)} according to its high uncertainty score. 
In this way, we can abstain from unreliable decisions of the pre-denoising model, which can mitigate the risk of false pseudo labels. Considering there might be multiple relations between an entity pair, we propose an instance-level UE method to capture uncertainty scores for overlapping relations. Moreover, we design a re-labeling strategy with dynamic class uncertainty thresholds by taking the DocRE long-tail problem into account to obtain high-quality DS data. With the proposed uncertainty guided label denoising mechanism, we design a multi-phase training strategy to further boost the performance of our final DocRE model.

The main contributions of our work are summarized as follows:
\begin{itemize}
\item We propose a document-level relation distant extraction framework with uncertainty guided label denoising, which greatly improves the label quality of DS data. 
\item We propose a novel instance-level uncertainty estimation method for overlapping relations to measure the reliability of instance-level pseudo labels. 
\item We design an iterative re-label strategy with dynamic class uncertainty thresholds for the problem of long-tail in DocRE to filter high uncertainty pseudo labels. 
\item The proposed framework achieves significant performance improvements over existing competitive baselines on two public datasets. Extensive experiments illustrate that the performance of baselines trained on our denoised DS (DDS) data is obviously improved. 
\end{itemize}

\section{Related Work}
\textbf{Sentence-level Relation Extraction.} 
Conventional works on RE mainly focus on sentence-level supervised relation extraction  \citep{zhou2016attention,guo2019attention,sun2022joint}. Although these models achieve great success in RE, they primarily rely on the large-scale human-annotated corpus that needs time-consuming labels.
Therefore, early works prefer to use extra data that are auto-labeled by distant supervision (DS) \citep{zeng2015distant,huang2021local,peng2022distantly,qin2018dsgan}. However, the noisy labels caused by distant supervision will influence the performance of these models. Thus, various works are proposed to select effective instances, separate noisy data, and enhance the robustness of models. Most of them tend to perform attention mechanism\citep{li2020self,yuan2019cross,han2018hierarchical}, negative training \citep{ma2021sent}, reinforcement learning\citep{feng2018reinforcement}, and soft-label strategies \citep{liu2017soft}. However, the above DS methods mainly focus on sentence-level RE, which can not be transferred to DocRE directly. 

\begin{figure*}
	\centering
	\includegraphics[width=0.62\textheight]{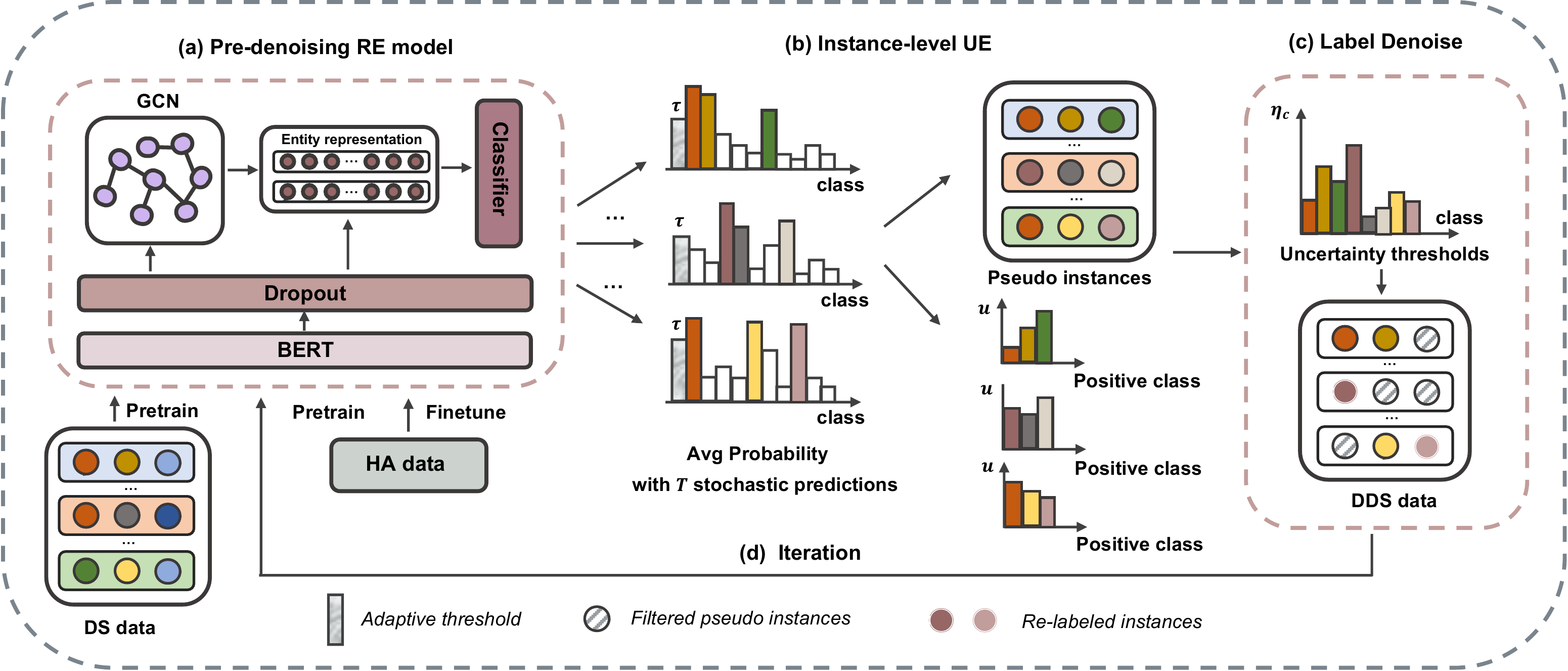}
	\caption{The overview of our \model \,  framework. It contains four key parts as follows: (a) Pre-denoising DocRE model; (b) instance-level UE of pseudo instances generated by Pre-denoising RE model; (c) Label denoising strategy to re-label with pseudo instances that contain low uncertainty scores; (d) Iterative training strategy for further re-label and improve the performance.}
	\label{fig:2}    

\end{figure*}

\textbf{Document-level Relation Extraction.} 
DocRE aims to extract relations between each entity pair expressed by multiple mentions across the sentences in a document. Different from the conventional sentence-level RE, DocRE needs the ability to reason relations in a more complex semantic scene. Existing approaches employ transformer-based models to extract contextual information for aggregating entity representations \citep{yao2019docred,huang2021three,zhou2021atlop,li2021mrn}. To further capture non-local syntactic and semantic structural information, some works construct document-level graphs and aggregate graph representations by Graph Neural Networks (GNN) \cite{sahu2019inter, wang2020global,eberts2021end,christopoulou2019connecting,nan2020reasoning,zeng2020double,wangx-ACL-2021-drn,zeng2021sire,sun2023document}. Recent works observe that utilizing large-scale auto-labeled data generated by distant supervision \citep{mintz2009distant} to pretrain the DocRE model can attain great performance improvements \cite{xu2021entity,zhou2022none,wang2022unified,hogan-etal-2022-fine}. Most of them directly utilize the DS data and ignore the accuracy of DS labels. To obtain high-quality DS data, several methods introduce re-label strategies based on the pre-denoising RE model trained on human-annotated data \citep{xiao2020denoising, tan2022KDDocRED}. However, these methods ignore the noise caused by pseudo labels. In this work, we introduce uncertainty estimation to determine the reliability of pseudo labels, which can reduce the noisy pseudo labels to further improve the quality of DS data.

\section{Methodology}
In this section, we introduce our proposed framework in detail. As shown in Figure \ref{fig:2}, our \model \,  contains four key components: (1) Training a document-level pre-denoising model by the original DS and human-annotated training data; (2) Estimating uncertainty scores of pseudo labels generated by the pre-denoising model; (3) Denoising the DS data by pseudo labels and uncertainty scores; (4) Leveraging a multi-phase training strategy to iteratively train the DocRE model by denoised DS (DDS) data and human-annotated data. 

\subsection{Problem Formulation}
 Given a document $D=\{s_i\}_{i=1}^{t}$, which is composed of $t$ sentences. Each document contains a set of entities $E=\{e_i\}_{i=1}^{q}$, where $q$ is the number of entities. An entity might be mentioned multiple times in a document, formulated as $e_i=\{m_j^i\}_{j=1}^{p_i}$, where $p_i$ is the number of times the entity $e_i$ is mentioned. The aim of the document-level relation extraction is to predict relation types between entities, formulated as $\{(e_i,e_j,r_k)|e_i,e_j \in E,r_k \in R\}$, where $R$ is the set of pre-defined relation types. In addition, there can be more than one relation type between a specific entity pair in a document. Thus, the DocRE task can be regarded as a multi-label classification task. In the document-level distant relation extraction setting, we have a clean human-annotated dataset and a distantly supervised dataset, while the quantity of DS data is significantly larger than the human-annotated data.

\subsection{Document-level Pre-denoising Model}
In order to alleviate the noisy label problem in the DS data, we construct a pre-denoising DocRE model to generate pseudo labels. As shown in Figure \ref{fig:2}(a), we adopt BERT \citep{devlin2019bert} to capture the contextual representation $\{z_i\}_{i=1}^{n}$, where $n$ is the number of tokens in a document. We also adopt a dropout layer to enhance the generalization ability of our DocRE model. 

To capture non-local dependency among entities, we construct a graph for each document. Specifically, we take all tokens in a document as nodes and connect them using the task-specific rules: (1) To capture the semantic information of mentions, tokens belonging to the same mention are connected. (2) To capture interactions between mentions, tokens of mentions belonging to the same entities are connected. (3) To capture the interactions of entities, tokens of entities that co-occur in a single sentence are connected. 

We construct the adjacency matrix according to the document graph and apply Graph Convolutional Networks (GCN) to capture graph representations $\{g_i\}_{i=1}^{n}$, which is formulated as follows:
\begin{equation}
g_{i}=\rho \left(\sum_{j=1}^{n}A_{ij}W z_j+b\right),
\end{equation}
where $W \in \mathbb{R}^{d\times d}$ and $b\in \mathbb{R}^{d} $ are trainable parameters. 
$z_j$ is the contextual representation of $j$-th token, which is introduced above.
$A_{ij}$ is the adjacency matrix of the document graph. 
$\rho $ is the activation function. To obtain the global representations $\{h_i\}_{i=1}^{n}$, we concatenate the contextual representations $\{z_i\}_{i=1}^{n}$ and graph representations $\{g_i\}_{i=1}^{n}$ as follows: 
\begin{equation}
    h_i=[z_i,g_i].
\end{equation}
Following \citet{zhou2021atlop}, we also apply logsumexp pooling \cite{jia2019document} to obtain entity representations $\{e_i\}_{i=1}^{q}$. Finally, group bilinear \citep{van2020uncertainty} is utilized to obtain the probabilities $\{p^c_{ij}\}_{c=1}^{N_c}$ of each class $c$ for the entity pair $(e_i,e_j)$ to predict relation types.

\subsection{Instance-level Uncertainty Estimation}
Uncertainty Estimation (UE) is a vital technology for misclassification detection \citep{vazhentsev2022uncertainty}, out-of-distribution instances detection \citep{van2020uncertainty}, and active learning \citep{burkhardt2018semisupervised}. In order to model the uncertainty in pre-denoising DocRE model, we introduce the Monte Carlo (MC) dropout \citep{gal2016dropout} technology into the DocRE task. As a popular UE technology, MC dropout is formally equivalent to approximate Bayesian inference in deep Gaussian processes \citep{gal2016dropout}. This method requires multiple stochastic forward-pass predictions with activated dropout to capture the model uncertainty. 

Previous works based on MC dropout \citep{gal2017deep,vazhentsev2022uncertainty} calculate the uncertainty score of the model prediction as follows: 
\begin{equation}
    u_s=\frac{1}{N_c}\sum_{c=1}^{N_c}(\frac{1}{N_t}\sum_{t=1}^{N_t}(p_t^c-\overline{p^c})^2),
\end{equation}
where $N_c$ is the number of the class number. $N_t$ is the number of stochastic forward passes. $p_t^c$ is the probability of the $c$-th class at $t$-th stochastic forward passes. $\overline{p^c}=\frac{1}{N_t}\sum_{t=1}^{N_t}p_t^c$ is the average probability of the $c$-th class. 

\begin{figure}
	\centering
	\includegraphics[width=0.3\textheight]{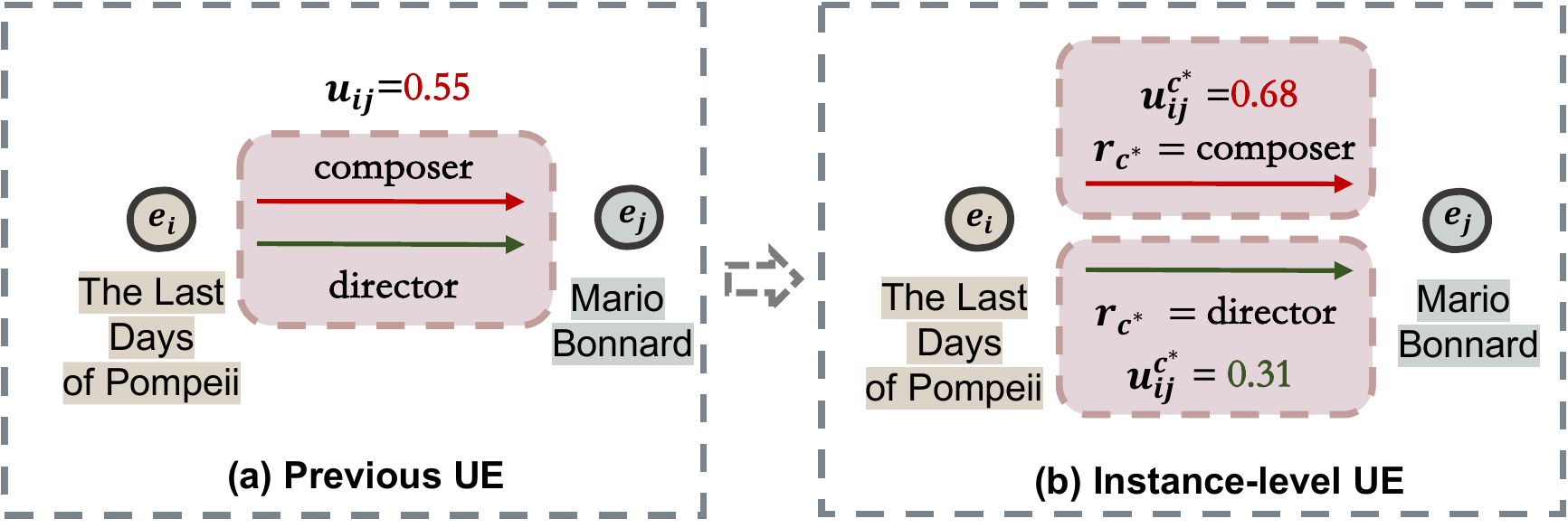}
	\caption{An example of our instance-level UE score for each predicted positive relation between an entity pair. We present two overlapping relations predicted by the pre-denoising model between an entity pair \textit{(The Last Days of Pompeii, Mario Bonnard)}.}
	\label{fig:3} 

\end{figure}

The above uncertainty estimation method provides one uncertainty score for each prediction. However, there exist multiple relations for one entity pair, which can be called overlapping relations. Intuitively, different overlapping relations should have their own uncertainty scores. As shown in Figure \ref{fig:3}(a), there are two different types of relations between an entity pair \textit{(The Last Days of Pompeii, Mario Bonnard)}. It is hard to separate the false positive pseudo label \textit{composer} and correct positive pseudo label \textit{director} by previous UE methods \citep{gal2017deep,vazhentsev2022uncertainty}. 

To solve this issue, we modify the estimation process to obtain the instance-level UE score for each positive pseudo label between an entity pair, which can be seen in Figure \ref{fig:3}(b). Inspired by ATLOP \citep{zhou2021atlop} that introduces a threshold class $\tilde{c}$ to separate positive and negative relation classes. We calculate the adaptive threshold score $\tau _{ij}$ for entity pair $(e_i,e_j)$ as follows: 
\begin{equation}
\tau _{ij}=\frac{1}{N_t}\sum_{t=1}^{N_t}p_{ijt}^{\tilde{c}},
\end{equation}
where $p_{ijt}^{\tilde{c}}$ is the probability of the threshold class for entity pair $(e_i,e_j)$ at $t$-th stochastic forward passes. $N_t$ is the number of stochastic forward passes. Then, we regard classes of which average probability $\overline{p^c_{ij}}=\frac{1}{N_t}\sum_{t=1}^{N_t}p_{ijt}^c$ are higher than the threshold $\tau _{ij}$ as positive classes. If all the class probabilities are lower than the probability of the threshold class, we regard ``NA (no relationship)'' as the relationship type for the entity pair. 
Then, we calculate the uncertainty score of each positive class for entity pair $(e_i,e_j)$ as follows:
\begin{equation}
u_{ij}^{c^*}=\frac{1}{N_t}\sum_{t=1}^{N_t}(p_{ijt}^{c^*}-\overline{p_{ij}^{c^*}})^2,c^*\in \{\overline{p^c_{ij}}>\tau_{ij}\},
\end{equation}
where $p_{ijt}^{c^*}$ is the probability of the positive class $c^*$ at $t$-th stochastic forward passes. $\overline{p_{ij}^{c^*}}=\frac{1}{N_t}\sum_{t=1}^{N_t}p_{ij}^{c^*}$ is the average probability of the positive class $c^*$. 

In this way, we can obtain each positive pseudo label with its uncertainty score between an entity pair, which is shown in Figure \ref{fig:2}(b). Each pseudo instance is formulated as $(e_i,e_j,r_{c^*},u^{c^*}_{ij})$.

\begin{figure}
	\centering\small
	\includegraphics[width=0.25\textheight]{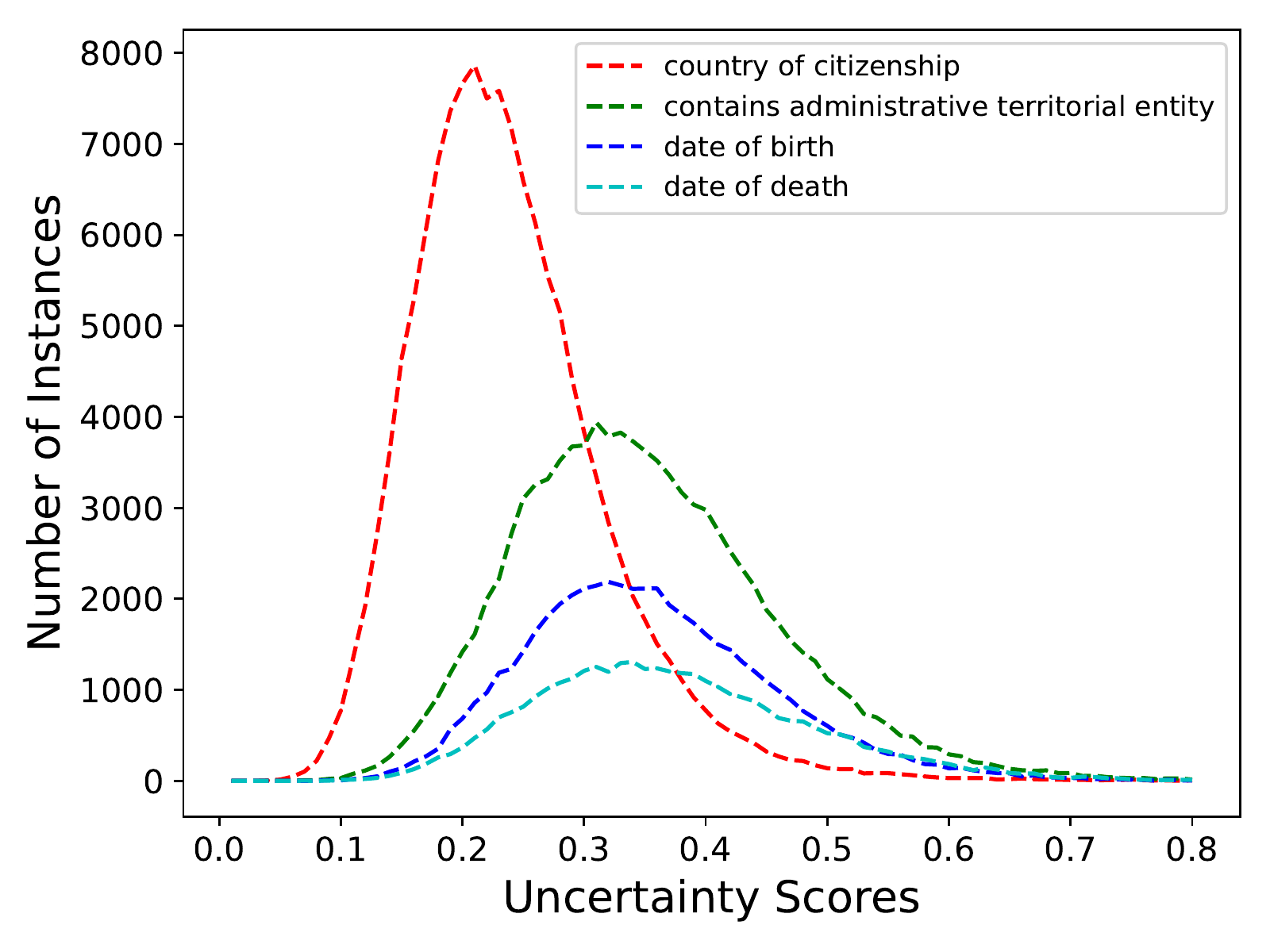}
	\caption{The distribution of UE scores of several relation types.}
	\label{fig:4}      

\end{figure}

\subsection{Uncertainty Guided Label Denoising}
After obtaining instance-level pseudo labels and their corresponding uncertainty scores, we re-label DS data by the proposed uncertainty-guided denoising strategy (Figure \ref{fig:2}(c)). We observe that the distribution of uncertainty scores for each relation class is obviously different, which is shown in Figure \ref{fig:4}. Moreover, it can be observed that frequent classes usually contain lower average uncertainty than long-tail classes. Therefore, considering the long-tail problem in the DocRE task, we propose dynamic class uncertainty thresholds to filter pseudo labels with high uncertainty. For each class of relation, the corresponding uncertainty threshold is calculated as follows: 
\begin{equation}
    \eta _{c^*}=\overline{u^{c^*}}+ \sqrt{\frac{1}{N^{\eta}_{c^*}-1}\sum_{l=1}^{N^{\eta}_{c^*}}(u_l^{c^*}-\overline{u^{c^*}})^2},
\end{equation}
where $u_l^{c^*}$ is the uncertainty score of the $l$-th pseudo instance of class $c^*$.
$\overline{u^{c^*}}=\frac{1}{N^{\eta}_{c^*}}\sum_{l=1}^{N^{\eta}_{c^*}}u_i^{c^*}$ is the average of uncertainty scores for class $c{^*}$ in all pseudo instances.  $N^{\eta}_{c^*}$ is the quantity of pseudo instances that belong to class $c{^*}$. 

In our re-label strategy (Figure \ref{fig:2}(c)), for each entity pair $(e_i,e_j)$, we replace the original DS label with the pseudo label $r_{c^*}$ that contain the lower uncertainty score $u^{c^*}_{ij}$ than its class uncertainty thresholds $\eta _{c^*}$.
In this way, we are able to reduce false positive pseudo labels and keep correct DS labels with high-quality pseudo labels. Besides, to reduce false negative pseudo labels, we keep the original DS positive labels where do not exist positive pseudo labels between an entity pair. 

\begin{algorithm}
\caption{Multi-phase Training Strategy}
\label{alg:1}
\textbf{Define:} 
Human-annotated training and test data: $HA$ and $DT$, DS data: $DS$, Iteration: $K$, DocRE Model: $M$, Pseudo labels with UE: $PU$, Denoised DS data: $DDS$, Relations: $TR$.

\begin{algorithmic}
\Require 
$DS$, $HA$, $K$, $M$.
\Ensure $K>0$.
\State $M_{pretrain} \gets Train \, (M, DS)$
\State $M_{finetune} \gets Train \, (M_{pretrain}, HA)$
\State \textbf{for} $i=1; \, i<=K; \, i++$ \textbf{do}
\State 1. $PU \gets Predict \, (M_{finetune}, DS)$
\State 2. $DDS \gets Denoise \, (DS,PU)$
\State 3. $M \gets Reinitialized \, (M_{finetune})$
\State 4. $M_{pretrain} \gets Train \, (M, DDS)$
\State 5. $M_{finetune} \gets Train \, (M_{pretrain}, HA)$
\State 6. $DS \gets DDS$

\State \textbf{end for}
\State  $TR \gets Predict \, (M_{finetune}, DT)$ \\
\Return $TR$
\end{algorithmic}
\end{algorithm}

\begin{table*}
\centering \small
\begin{tabular}{l c c c c c c c c}
\toprule
\multirow{3}*{\textbf{Model}} & \multicolumn{4}{c}{\textbf{DocRED}} & \multicolumn{4}{c}{\textbf{Re-DocRED}} \\
\cmidrule(r){2-5}
\cmidrule(r){6-9}
 &\multicolumn{2}{c}{\textbf{Dev}} & \multicolumn{2}{c}{\textbf{Test}}&\multicolumn{2}{c}{\textbf{Dev}} & \multicolumn{2}{c}{\textbf{Test}}\\
\cmidrule(r){2-3}
\cmidrule(r){4-5}
\cmidrule(r){6-7}
\cmidrule(r){8-9}
 & $F_1$ & Ign $F_1$ & $F_1$ & Ign $F_1$ & $F_1$ & Ign $F_1$ & $F_1$ & Ign $F_1$ \\
\midrule

ATLOP \citep{zhou2021atlop} & 63.42 & 61.57 & 63.48 & 61.43 & 74.34 &  73.62 & 74.23 & 73.53 \\
DocuNet \citep{zhang2021document} & 64.35 & 62.66 & 64.00 & 61.93 & 76.22 & 75.50 & 75.35 & 74.61\\
NCRL \citep{zhou2022none} & 63.87 & 61.65 & 63.45 & 60.98 & 75.85 & 74.91 & 75.90 & 75.00\\
SSR-PU \citep{wang2022unified} & 63.00  & 60.43 & 62.66 & 59.80 & 76.83& 75.57 &76.23 & 74.96  \\
KD-NA* \citep{tan2022KDDocRED} & 64.17 & 62.18 & 64.12 & 61.77 & 76.14 & 75.25 & 76.00 & 75.12 \\
KD-DocRE* \citep{tan2022KDDocRED} & 64.81 & 62.62 & 64.76 & 62.56 &  76.47 & 75.30 & 76.14 & 74.97 \\
\midrule
\textbf{\model \, (Ours)} & \textbf{65.71} & \textbf{63.62} & \textbf{65.58} & \textbf{63.26} & \textbf{78.28} & \textbf{77.32} & \textbf{78.14} & \textbf{77.24}\\
\bottomrule
\end{tabular}
\caption{\label{tab:1}
Experimental results on public datasets: DocRED and Re-DocRED dataset. The baselines are trained with both DS data and human-annotated training data. The test results of DocRED are obtained from the leaderboard submission. Results of DocRED with * are from \citet{tan2022KDDocRED}
}

\end{table*}

\subsection{Multi-phase Training Strategy}
In order to take full advantage of the DS data for further boosting the performance of the DocRE model, we design a multi-phase training strategy to iteratively re-label the DS data, which is shown in Algorithm \ref{alg:1}. We introduce the overall process as follows. (1) We train the initial pre-denoising RE model with the original DS data and human-annotated data. (2) We leverage the pre-denoising RE model to generate pseudo instances with uncertainty scores on DS data. (3) We perform a re-label strategy to obtain denoised distantly supervised (DDS) data. (4) We re-initialize and train the pre-denoising DocRE with DDS data and human-annotated data to boost performance. We iterate the above (2), (3), and (4) phases until we obtain the best DocRE model.

\section{Experiments}
\subsection{Dataset and Settings}
\textbf{Dataset.} DocRED \citep{yao2019docred} is a popular DocRE dataset with 96 pre-defined relation types, which is constructed from Wikipedia and Wikidata. It provides a distant-supervised dataset with 101873 documents and a large-scale human-annotated dataset with 5053 documents. Re-DocRED is a high-quality revised version of human-annotated documents of DocRED, which is provided by \citet{tan2022revisiting} recently. Re-DocRED contains 3053, 500, and 500 documents for training, development, and testing. See Appendix \ref{adx:1} for more details.

\textbf{Settings.} Following previous works 
\citep{zhou2021atlop,tan2022KDDocRED}, we adopt BERT$_{base}$ \citep{devlin2019bert} as the context encoder. We use AdamW \citep{loshchilov2017decoupled} as the optimizer. We set the learning rate to 3e-5. We apply warmup for the initial 6\% steps. We set the batch size to 8 for both the training and test process. The rate of the dropout is 0.25. All hyper-parameters are tuned on the development set. The experiments are conducted on a single NVIDIA RTX A6000-48G GPU. DocRED and RE-DocRED both contain 3053 human-annotated training documents for fine-tuning and 101873 distantly supervised training documents for pretraining. Thus, for each dataset, our framework takes about 55 hours and consumes about 23G GPU memory for training. Following \citet{yao2019docred}, we use $F_1$ and $IgnF_1$ as the evaluation metrics. The $IgnF_1$ represents $F_1$ scores, which excludes the relational facts shared by the human-annotated training set.  

\subsection{Compared Methods}
We compare our \model \,  with several strong baselines that are trained on both DS and human-annotated data. \textbf{ATLOP} \citep{zhou2021atlop} utilizes an adaptive thresholding loss to solve the overlapping relation problem, and adopts a localized context pooling to aggregate entity representations. \textbf{DocuNet} \citep{zhang2021document} regards the DocRE task as a semantic segmentation task that provides a new view to extract document-level relations. \textbf{NCRL} \citep{zhou2022none} uses a multi-label loss that prefers large label margins between the ``NA'' relation class and each predefined class. \textbf{SSR-PU} \citep{wang2022unified} is a positive-unlabeled learning framework, which adapts DocRE with incomplete labeling. \textbf{KD-DocRE} \citep{tan2022KDDocRED} attempts to overcome the differences between human-annotated and DS data by knowledge distillation. They also provide the \textbf{KD-NA} \citep{tan2022KDDocRED}, which is pretrained by DS data first and then fine-tuned by human-annotated data.

\begin{table*}
\centering \small
\begin{tabular}{l c c c c c c c c c c c c}
\toprule
\multirow{3}*{\textbf{Model}} & \multicolumn{4}{c}{\textbf{Origin}} & \multicolumn{4}{c}{\textbf{After Denoising}} &\multicolumn{2}{c}{\multirow{3}*{\textbf{Improvement}}} \\
\cmidrule(r){2-5}
\cmidrule(r){6-9}

 &\multicolumn{2}{c}{\textbf{Dev}} & \multicolumn{2}{c}{\textbf{Test}}&\multicolumn{2}{c}{\textbf{Dev}} & \multicolumn{2}{c}{\textbf{Test}}\\
\cmidrule(r){2-3}
\cmidrule(r){4-5}
\cmidrule(r){6-7}
\cmidrule(r){8-9}
\cmidrule(r){10-11}

 & $F_1$ & Ign $F_1$ & $F_1$ & Ign $F_1$ & $F_1$ & Ign $F_1$ & $F_1$ & Ign $F_1$ & $\Delta F_1$ &$\Delta$Ign $F_1$ \\
\midrule
\multicolumn{11}{l}{\textbf{+DocRED}}\\
ATLOP  & 54.38 & 51.62 & 53.10 &50.01 & 59.00 & 56.35 & 58.34 & 55.35 & +5.24 & +5.34  \\
DocuNet  & 53.79 & 50.91 & 52.96 & 49.69& 59.03 & 56.17 & 58.05 & 54.85 & +5.09& +5.16\\
NCRL & 54.53 & 51.66 & 53.26 & 50.03& 59.39 & 56.71 & 58.50&55.46& +5.24 & +5.43\\
KD-NA & 54.02 & 50.94 & 54.10 & 50.65& 58.39 & 55.31 & 58.20 & 54.79& +4.10 & +4.14\\
\textbf{\model \, (Ours)} & 54.74 & 51.91 & 54.47 & 51.27 & 59.75 & 56.84 & 58.92 & 55.67 & +4.45 & +4.40\\
\midrule
\multicolumn{11}{l}{\textbf{+RE-DocRED}}\\
ATLOP  & 43.48 & 42.69 & 42.59 & 41.77 & 75.99 & 74.86 & 75.29 & 74.16& +32.70 & +32.39\\
DocuNet  & 44.22 & 43.38 & 43.89 & 43.02 & 76.38 & 75.18 & 75.64 & 74.44 & +31.75  & +31.42 \\
NCRL &  44.71 & 43.87 & 44.09 & 43.23 & 76.39 & 75.19 & 75.69 & 74.50 & +31.60 & +31.27\\
KD-NA  & 45.55 & 44.58 & 45.38 & 44.41 & 76.11 & 74.78 & 75.37 & 74.00 & +29.99 & +29.59 \\
\textbf{\model \, (Ours)} &  45.56 &  44.71 & 44.76 & 43.94 & 76.47 & 75.24 & 75.57 & 74.32 & +30.81 & +30.38\\

\bottomrule
\end{tabular}
\caption{\label{tab:2}
Experimental results of DocRE baselines trained on original DS data and our denoised DS (DDS) data. 
}

\end{table*}

\subsection{Experimental Results}
We compare our \model \, framework with the above baselines, which are also based on BERT$_{base}$ \citep{devlin2019bert} and trained on both DS data and human-annotated data. As shown in Table \ref{tab:1}, our framework \model \, outperforms the previous baselines on both DocRED and RE-DocRED datasets. Specifically, our \model \,  achieves 65.71 $F_1$ and 78.14 $F_1$ on the test set of DocRED  and RE-DocRED datasets, respectively. Our \model \, outperforms the KD-DocRE \citep{tan2022KDDocRED} that leverages knowledge distillation to denoise by 2.00 $F_1$ and 0.82 $F_1$ on the test set of RE-DocRE and DocRED datasets. Moreover, our \model \,  significantly outperforms the latest strong baseline SSR-PU \cite{wang2022unified} by 1.91 $F_1$ and 2.28 Ign $F_1$ on the Re-DocRED dataset. This suggests the effectiveness of our uncertainty-guided denoise strategy. 

Besides, we observe that improvements on the RE-DocRED dataset are obviously higher than DocRED dataset, which can be caused by the following: 1) The RE-DocRED dataset is a revised version of the DocRED dataset by adding more positive instances. It alleviates the imbalance problem of positive and negative instances. 2) The pre-denoising model trained on RE-DocRED achieves a higher ability to discover relations, which will enhance the denoise process.

\section{Analysis and Discussion}
In this section, we conduct extensive experiments to further analyze the effectiveness of our proposed denoising strategy and instance-level UE method. We also conduct the ablation study to discuss the contribution of each component of the framework. 

\begin{figure}

  \centering
\subfigure[Frequent relation.]{
   \label{fig:5_1}
  \includegraphics[scale = 0.22]{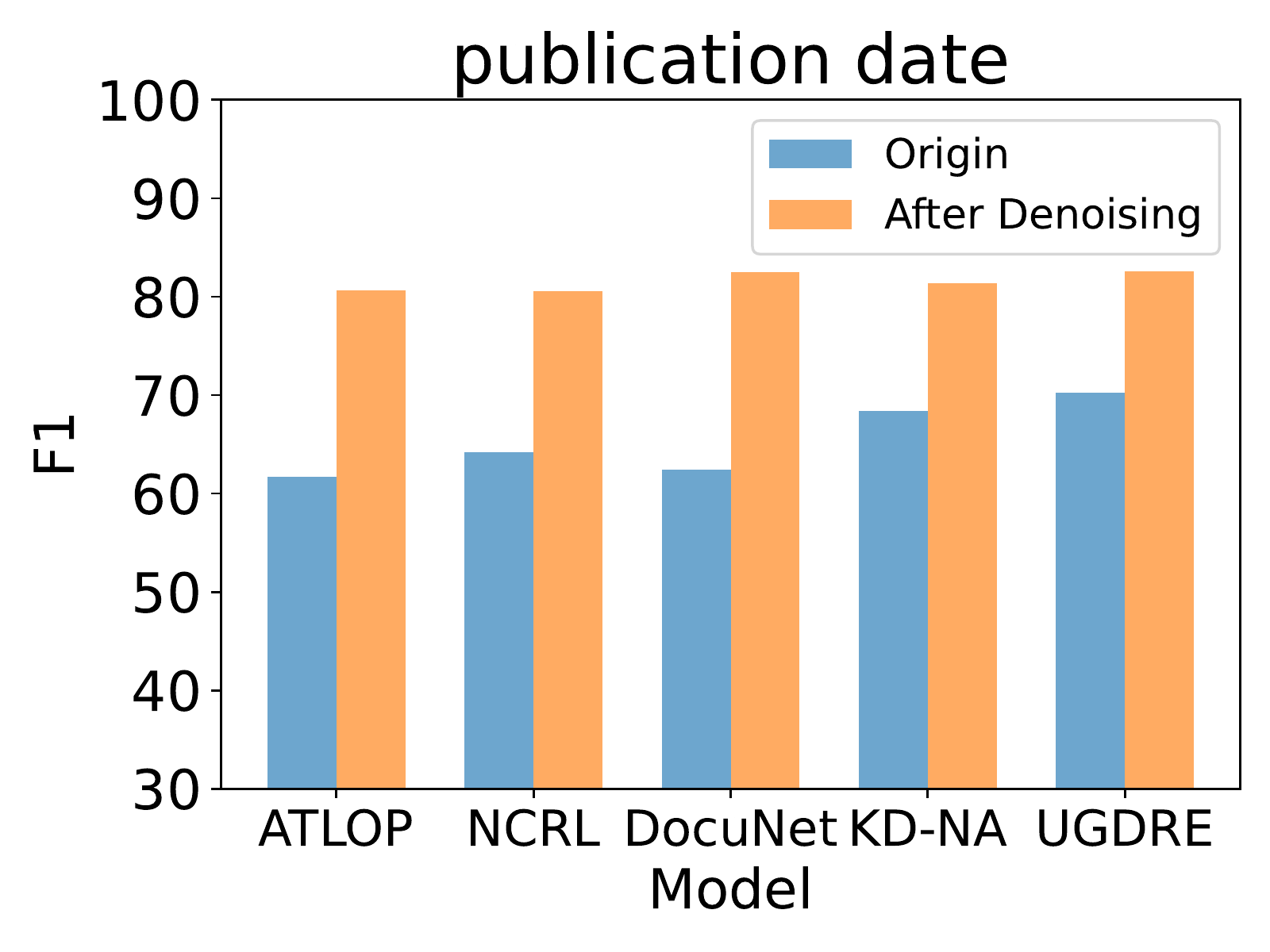}}
\subfigure[Long-tail relation.]{
  \label{fig:5_2}
  \includegraphics[scale = 0.22]{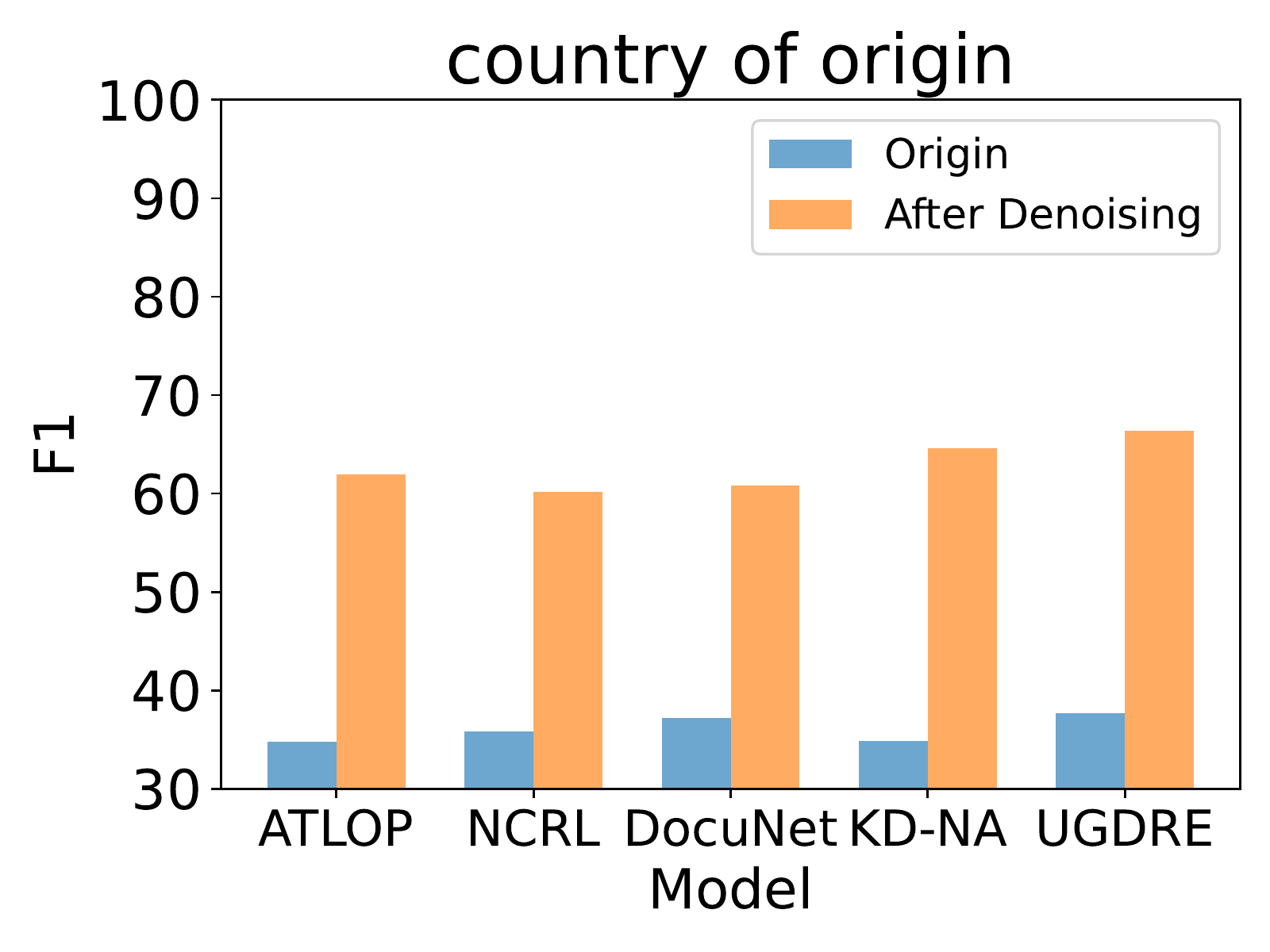}}
\caption{Experiment results for the frequent relation type \textit{publication date} and long-tail relation type \textit{country of origin} on the RE-DocRED dataset.} 
  \label{fig:5} 

\end{figure}

\subsection{Effectiveness of the Denoising Strategy}
In order to intuitively demonstrate the effectiveness of our uncertainty-guided denoising strategy. We present experimental results of several DocRE baselines only trained on original DS data and our denoised DS (DDS) data. As shown in Table \ref{tab:2}, we can observe that all baselines trained on our DDS data obtain significant performance improvements on both DocRED and RE-DocRED. In contrast to the original DS data, the performance of baselines trained on our DDS data increases more than 4 $F_1$ and 29 $F_1$ on the test set of the DocRED and RE-DocRED datasets. This further demonstrates the effectiveness of our uncertainty guided denoising strategy. 

We observe that when training on original DS data, the performance of baselines on the RE-DocRED dataset is obviously lower than the DocRED dataset. This is because there are more positive instances in the RE-DocRED dataset than in the DocRED dataset, which makes the noise problem more obvious. Thus, the performance improvement of models trained on our denoised data will also be more obvious. The performance improvements of baselines that are fine-tuned on human-annotated data can be seen in Appendix \ref{adx:2}.

\begin{figure*}
	\centering
	\includegraphics[width=0.62\textheight]{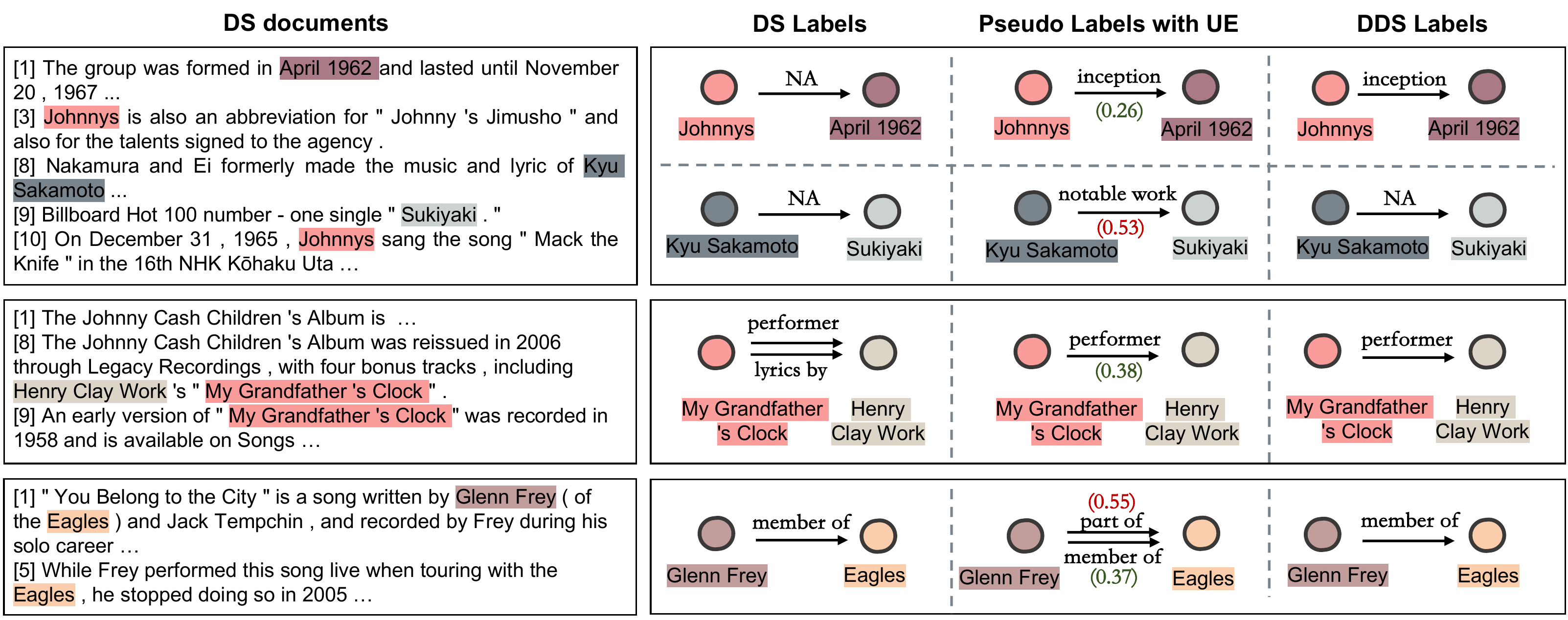}
	\caption{Case Study. We present several samples from the DS data, which contain original DS labels, pseudo labels with our proposed instance-level UE scores, and denoised distantly supervised (DDS) labels. We mark the UE scores that exceed their class uncertainty thresholds with red color.}
	\label{fig:6}      

\end{figure*}

In addition, we also evaluate the performance of the above models on each relation type. As shown in Figure \ref{fig:5}, the performance improvement of the long-tail relation type \textit{country of origin} is obviously higher than the frequent relation type \textit{publication date} after training on our denoised data. This indicates the effectiveness of our dynamic class uncertainty thresholds designed for the long-tail problem in DocRE task.

\begin{table}

\centering \small
\begin{tabular}{l c c c c c c }
\toprule
\multirow{2}*{\textbf{Model}}&\multicolumn{2}{c}{\textbf{Dev}} & \multicolumn{2}{c}{\textbf{Test}}  \\
\cmidrule(r){2-3}
\cmidrule(r){4-5}

 & $F_1$ & Ign $F_1$ & $F_1$ & Ign $F_1$  \\
\midrule

SR & 73.41 &72.43 & 72.40 & 71.39\\
Entropy &  74.42& 73.27 & 73.49 & 72.31\\
PV Dropout & 75.54 & 74.41 & 74.87 & 73.71\\
\textbf{\model } &\textbf{76.47} & \textbf{75.24} & \textbf{75.57} & \textbf{74.32} \\

\bottomrule
\end{tabular}
\caption{\label{tab:3}
Comparison of uncertainty estimation methods on Re-DocRED dataset.
}

\end{table}

\subsection{Effectiveness of Instance-level Uncertainty Estimation}
 We compare our proposed instance-level UE with existing popular UE technologies \citep{vazhentsev2022uncertainty} as follows: 1) Softmax Response (SR); 2) Entropy; 3) Probability Variance (PV) with MC dropout. The performance of the DocRE model trained on denoised DS data that performed different UE technology is shown in Table \ref{tab:3}. It can be observed that the DocRE model based on our instance-level UE outperforms SR, entropy, and PV dropout based methods on the test set of the RE-DocRED dataset. This is because our instance-level UE provides specific UE scores for different overlapping relations, which enables our downstream uncertainty guided relabel strategy to separate the false pseudo label from the overlapping relations. The experimental results also demonstrate the effectiveness of our proposed instance-level UE method.

\subsection{Case Study}
We present several samples of DS data that are denoised by our \model \, framework in Figure \ref{fig:6}. It can be observed that our framework denoises the DS data by 1) adding the extra correct positive instance, such as \textit{(Johnnys, April 1962, inception)}; 2) Removing false DS instances, such as \textit{(My Grandfather's Clock, Henry Clay Work, lyrics by)}. Moreover, we also present pseudo labels with our instance-level UE scores to show the process of re-relabel strategy. As shown in the second and fourth samples of Figure \ref{fig:6}, our framework is able to reduce the false pseudo labels by their high uncertainty scores, such as \textit{(Kyu Sakamoto, Sukiyaki, notable work)} and \textit{(Glenn Frey, Eagles, member of)}. 

\begin{table}

\centering\small
\begin{tabular}{l c c c c c c }
\toprule
\multirow{2}*{\textbf{Model}}&\multicolumn{2}{c}{\textbf{Dev}} & \multicolumn{2}{c}{\textbf{Test}}  \\
\cmidrule(r){2-3}
\cmidrule(r){4-5}

 & $F_1$ & Ign $F_1$ & $F_1$ & Ign $F_1$  \\
\midrule
\textbf{\model } &78.28 & 77.32 & 78.14 & 77.24 \\
w/o Pretrain &   74.25&  73.36 & 74.10 & 73.21\\
w/o DDS & 76.91 & 76.00 & 76.16 & 75.23\\
w/o UE & 77.66 & 76.80& 76.84 & 75.99\\

\bottomrule
\end{tabular}
\caption{\label{tab:4}
Ablation study on the RE-DocRED dataset.
}

\end{table}

\subsection{Ablation Study}
To analyze the effectiveness of each component in our \model \, framework, we conduct the ablation study by removing different components. As shown in Table \ref{tab:3}, the performance decreases as removing each component, which demonstrates the effectiveness of our framework. When we remove the pretrain mechanism with DS data, the DocRE model trained by merely human-annotated data achieves 74.10 $F_1$ and 73.21 Ign $F_1$ on the test set of RE-DocRED dataset. This drop demonstrates that leveraging DS data can enhance the performance of the DocRE model. Removing the denoised distantly supervised (DDS) data leads to a 1.98 and 2.01 drop in terms of $F_1$ and Ign $F_1$ on the test set of RE-DocRED dataset. This indicates the significant effect of our uncertainty guided label denoising strategy. Our \model \, framework is also effective on sentence-level distant RE, which can be seen in Appendix \ref{adx:3}.

\begin{figure}
	\centering\small
	\includegraphics[width=0.25\textheight]{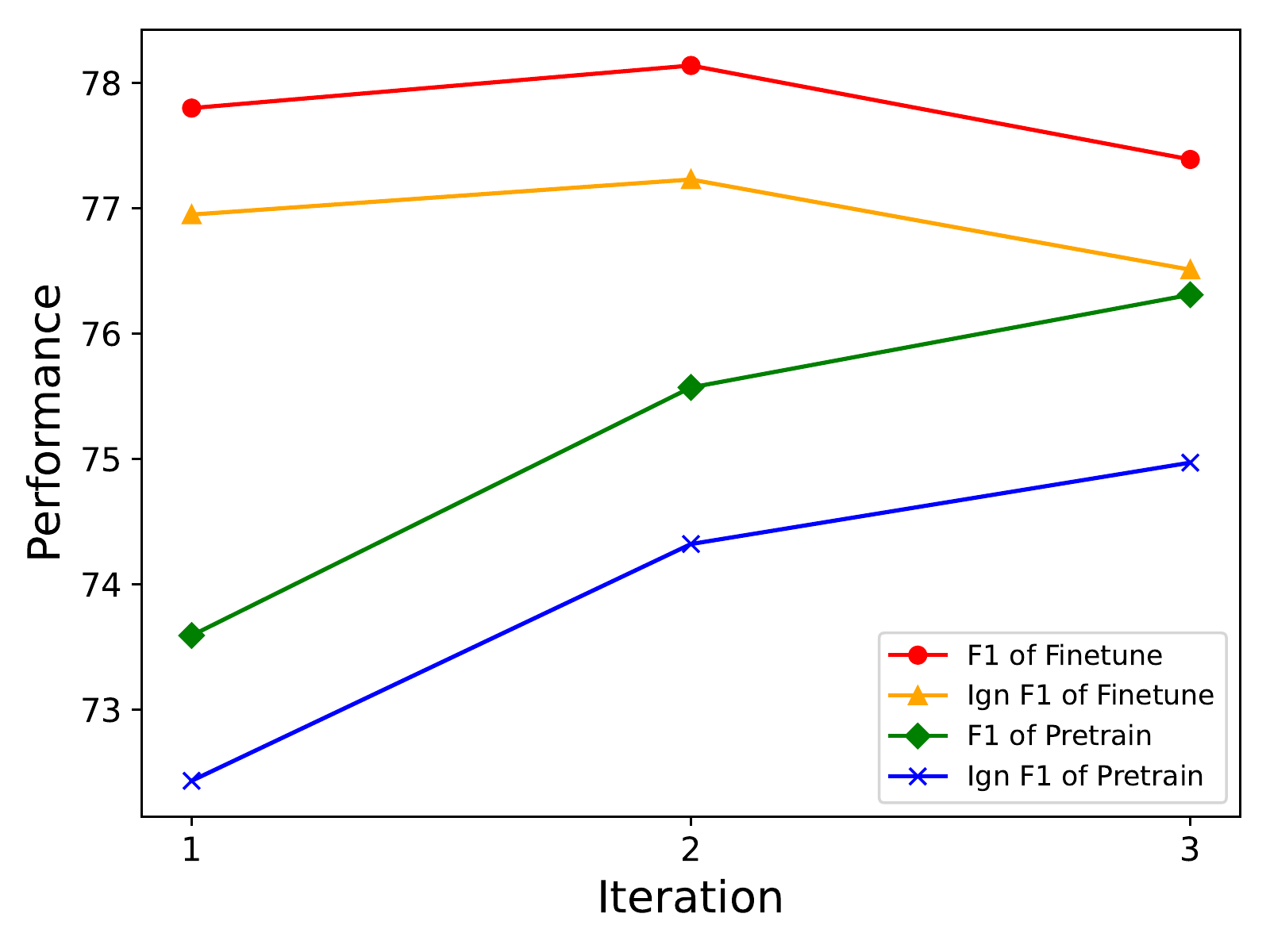}
	\caption{Performance of the model under different iterations on the test set of RE-DocRED.}
	\label{fig:7}      

\end{figure}
As shown in Figure \ref{fig:7}, we also present the performance of each iteration of the model that is pretrained on DDS and fine-tuned on human-annotated data. We can observe that the final model performance achieves the best by the second iteration of Algorithm \ref{alg:1}, which proves the effectiveness of our multi-phase training strategy. Moreover, the removal of our instance-level uncertainty estimation also causes an obvious drop, which illustrates the importance of estimating uncertainty in our framework.

\section{Conclusion}
In this paper, we propose a \textbf{D}ocument-level distant \textbf{R}elation \textbf{E}xtraction framework with \textbf{U}ncertainty \textbf{G}uided label denoising, \model. Specifically, we propose instance-level uncertainty estimation to measure the reliability of pseudo labels. Considering the long-tail problem, we design dynamic class uncertainty thresholds to filter high-uncertainty pseudo labels. Our proposed uncertainty guided denoising strategy can improve the quality of DS data. Experimental results demonstrate that our \model \,  outperforms competitive baselines. Moreover, extensive experiments verify the effectiveness of our label denoising. There are various challenges in DocRE worth exploring, one is to research the low-resource relation extraction. 

\section*{Limitations}
In this section, we discuss the limitations of our proposed framework. Our \model \, can reduce the false positive pseudo label by estimating the uncertainty of the model prediction. However, it is difficult to reduce the false negative pseudo labels by uncertainty estimation. Our framework also relies on human-annotated data to train the pre-denoising model, which causes the sensitivity of our framework to the quality of human-annotated data. Thus, the improvements of models that continue training on the DocRED dataset are not as well as on the RE-DocRED dataset. Moreover, iterative training introduces additional computing overhead, which makes the training process time-consuming. 

\section*{Acknowledgements}
Thanks to all co-authors for their hard work. The work is supported by the Chinese Scholarship Council, the National Program on Key Research Project of China (Project no. 2020XXXXXX6404), the Ministry of Education, Singapore, under its AcRF Tier-2 grant (Project no. T2MOE2008, and Grantor reference no. MOE-T2EP20220-0017), and A*STAR under its RIE 2020 AME programmatic grant (project reference no. RGAST2003). Any opinions, findings, and conclusions, or recommendations expressed in this material are those of the author(s) and do not reflect the views of the Ministry of Education, Singapore.

\bibliography{anthology,custom}
\bibliographystyle{acl_natbib}

\appendix

\section{Appendix}
\subsection{Statistics of Datasets} 
\label{adx:1}
We present statistics of the public DocRE datasets, including DocRED \citep{yao2019docred}, RE-DocRED \citep{tan2022revisiting}, and the distantly supervised data provided by \citet{yao2019docred}.

\begin{table}[H]
\centering \small
\begin{tabular}{l c c c }
\toprule

 Dataset&\# Document & Avg. \# Instance \\
\midrule
DocRED-Train & 3,053 &12.5 \\
DocRED-Dev & 1,000 & 12.3 \\
DocRED-Test & 1,000 &12.8 \\
Re-DocRED-Train & 3,053 &28.1 \\
Re-DocRED-Dev & 500 & 34.6  \\
Re-DocRED-Test & 500 &34.9 \\
Distantly Supervised &101,873 &14.8  \\

\bottomrule
\end{tabular}
\caption{\label{tab:5}
Statistics of the Re-DocRED, DocRED, and distantly supervised dataset.
}
\end{table}

\subsection{Results of Baselines with Fine-tuning}
\label{adx:2}
We present the results of baseline models that are pretrained on our denoised data and fine-tuned on the human-annotated data of the RE-DocRED dataset. As shown in Table \ref{tab:6},  the final performance of most baseline models that are pretrained on our denoised data is significantly improved.

\begin{table*}
\centering \small
\begin{tabular}{l c c c c c c c c c c c c}
\toprule
\multirow{3}*{\textbf{Model}} & \multicolumn{4}{c}{\textbf{DS}} & \multicolumn{4}{c}{\textbf{After Denoising}} &\multicolumn{2}{c}{\multirow{3}*{\textbf{Improvement}}} \\
\cmidrule(r){2-5}
\cmidrule(r){6-9}

 &\multicolumn{2}{c}{\textbf{Dev}} & \multicolumn{2}{c}{\textbf{Test}}&\multicolumn{2}{c}{\textbf{Dev}} & \multicolumn{2}{c}{\textbf{Test}}\\
\cmidrule(r){2-3}
\cmidrule(r){4-5}
\cmidrule(r){6-7}
\cmidrule(r){8-9}
\cmidrule(r){10-11}

 & $F_1$ & Ign $F_1$ & $F_1$ & Ign $F_1$ & $F_1$ & Ign $F_1$ & $F_1$ & Ign $F_1$ & $\Delta F_1$ &$\Delta$Ign $F_1$ \\
\midrule

ATLOP  & 74.34 &  73.62 & 74.23 & 73.53 & 77.30 & 76.63 & 76.95 & 76.28 & 2.72 &2.75 \\
DocuNet & 76.22 & 75.50 & 75.35 & 74.61 & 77.69 & 76.90 & 77.72 & 76.97 & 2.37 & 2.36\\
NCRL &  75.85 & 74.91 & 75.90 & 75.00 & 77.71 & 76.84 & 76.78 & 75.92  & 0.88 & 0.92 \\
KD-NA  & 76.14 & 75.25 & 76.00 & 75.12 & 78.16 & 77.23 &77.73 & 76.86 & 1.73 &  1.74\\
\textbf{\model \, (Ours) } & 76.91 & 76.00 & 76.16 & 75.23 & 78.28 & 77.32 & 78.14 & 77.24 & 1.98 & 2.01 \\
\bottomrule
\end{tabular}
\caption{\label{tab:6}
Experimental results of baselines fine-tuned on human-annotated training data of RE-DocRED dataset, which are pretrained on original DS data and our denoised DS data.
}
\end{table*}

\begin{table}

\centering\small
\begin{tabular}{l c c c c c c }
\toprule
\multirow{2}*{\textbf{Model}}&\multicolumn{2}{c}{\textbf{Dev}} & \multicolumn{2}{c}{\textbf{Test}}  \\
\cmidrule(r){2-3}
\cmidrule(r){4-5}

 & $F_1$ & Ign $F_1$ & $F_1$ & Ign $F_1$  \\
\midrule
\textbf{\model -SRE } &79.00 & 78.38 & 78.52 & 77.94 \\
w/o Pretrain &   76.88&  76.28 & 76.41 & 75.86\\
w/o DDS & 78.08 & 77.47 & 77.68 & 77.14 \\

\bottomrule
\end{tabular}
\caption{\label{tab:7}
Experimental results of our framework on sentence-level relation extraction task.
}

\end{table}

\subsection{Sentence-level Relation Extraction}
\label{adx:3}
Our framework can also be applied to the sentence-level relation extraction task. We reconstruct a sentence-level relation extraction dataset from the distantly supervised training data and RE-DocRED datasets. The sentence-level RE dataset contains 231,107 DS training data, 10,033 human-annotated training data, 1,862 human-annotated development data, and 1,794 human-annotated test data. 
We perform our \model \, on the sentence-level RE task in the same training way. 
As shown in Table \ref{tab:7}, the performance of the final sentence-level RE model UGDRE-SRE that pretrained on the DDS data is also improved.

\end{document}